\documentclass{article}

\usepackage{microtype}
\usepackage{graphicx}
\usepackage{subfigure}
\usepackage{booktabs} 
\usepackage{balance}

\usepackage{hyperref}

\usepackage[accepted]{icml2023}

\usepackage{amsmath}
\usepackage{amssymb}
\usepackage{mathtools}
\usepackage{amsthm}

\usepackage[capitalize,noabbrev]{cleveref}

\theoremstyle{plain}

\theoremstyle{definition}

\theoremstyle{remark}

\usepackage[textsize=tiny]{todonotes}

\icmltitlerunning{Efficient Rotation Invariance in Deep Neural Networks through Artificial Mental Rotation}

\begin{document}

\twocolumn[
\icmltitle{Efficient Rotation Invariance in Deep Neural Networks through Artificial Mental Rotation}



\icmlsetsymbol{equal}{*}

\begin{icmlauthorlist}
\icmlauthor{Lukas Tuggener}{CAI,USI}
\icmlauthor{Thilo Stadelmann}{CAI,VEN}
\icmlauthor{Jürgen Schmidhuber}{USI,IDSIA,KAUST}

\end{icmlauthorlist}

\icmlaffiliation{CAI}{ZHAW Centre for Artificial Intelligence, Winterthur, Switzerland}
\icmlaffiliation{USI}{University of Lugano, Switzerland}
\icmlaffiliation{IDSIA}{The Swiss AI Lab IDSIA, Switzerland}
\icmlaffiliation{KAUST}{AI Initiative, King Abdullah University of Science and Technology (KAUST), Saudi Arabia}
\icmlaffiliation{VEN}{European Centre for Living Technology, Venice, Italy}

\icmlcorrespondingauthor{Lukas Tuggener}{tugg@zhaw.ch}

\icmlkeywords{Machine Learning, ICML}
\vskip 0.3in
]



\printAffiliationsAndNotice{} 

\begin{abstract}
Humans and animals recognize objects irrespective of the beholder's point of view, which may drastically change their appearances. Artificial pattern recognizers also strive to achieve this, e.g., through translational invariance in convolutional neural networks (CNNs).
However, both CNNs and vision transformers (ViTs) perform very poorly on rotated inputs. Here we present \emph{artificial mental rotation} (AMR), a novel deep learning paradigm for dealing with in-plane rotations inspired by the neuro-psychological concept of mental rotation. Our simple AMR implementation works with all common CNN and ViT architectures. We test it on ImageNet, Stanford Cars, and Oxford Pet. With a top-1 error (averaged across datasets and architectures) of $0.743$, AMR outperforms the current state of the art (rotational data augmentation, average top-1 error of $0.626$) by $19\%$. We also easily transfer a trained AMR module to a downstream task to improve the performance of a pre-trained semantic segmentation model on rotated CoCo from $32.7$ to $55.2$ IoU.
\end{abstract}

\section{Introduction}

\begin{figure}[ht!]
\centering
\includegraphics[width=0.20\textwidth]{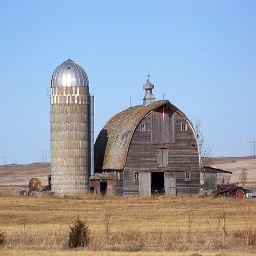}
\includegraphics[width=0.20\textwidth]{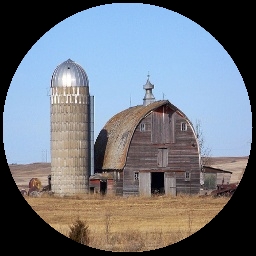} \\
\includegraphics[width=0.45\textwidth]{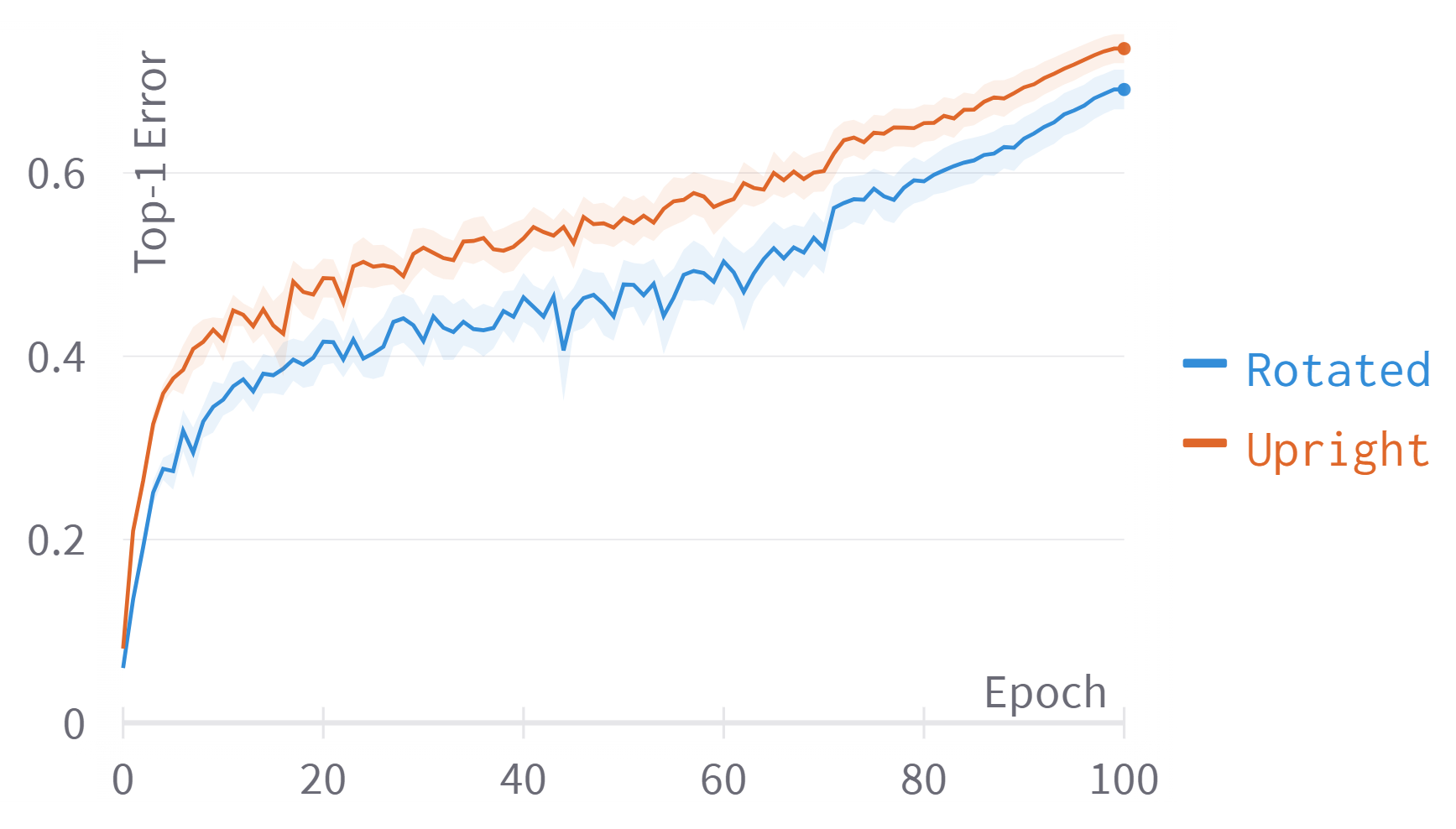}
\label{fig_data_aug_training_time}
\caption{An example image from ImageNet (top left) and a version of the same image with its corners masked, to allow for non-obvious and reversible rotations (top right). The top-1 classification error on ImageNet aggregated across $8$ CNN architectures for upright (orange) and rotated (blue) training, revealing a training slow-down for rotated inputs (bottom).}
\label{fig_mask}
\end{figure}

Natural vision systems in humans and animals are able to recognize objects irrespective of transformations such as rotations and translations as well as the observer's point of view, all of which can have a tremendous impact on the appearance of said object. This is a highly desirable property for all vision systems, especially if they are to be deployed in a real-world setting \cite{stadelmann2018deep, stadelmann2019beyond}.
Both CNNs \cite{fukushima1983neocognitron} and transformers \cite{DBLP:conf/nips/VaswaniSPUJGKP17} (a.k.a. neural fast weight programmers \cite{DBLP:journals/neco/Schmidhuber92, schlag2021linear}) inherently integrate translational invariance into their design. For rotations, this is not the case and both methods perform very poorly when facing inputs at an unusual angle. This can be exploited for adversarial attacks \cite{DBLP:conf/icml/EngstromTTSM19} and causes serious issues in applications where rotated inputs are common. There are currently two main research avenues trying to alleviate this issue: 

One is focused on building architectures that incorporate rotational invariance or sometimes equivariance directly into the neural network design. \citet{DBLP:conf/icml/CohenW16} introduced Group Equivariant Convolutional Neural Networks (G-CNNs) which are equivariant to a discrete symmetry group of rotations. \citet{DBLP:conf/iccv/GonzalezVKT17} on the other hand proposed to rotate the convolutional filters instead of the representations. A common drawback among rotation equivariant neural networks is that the memory footprint grows linearly with the angular resolution, severely limiting their practical use.

Another much more widely used approach is based on input data augmentation. The data is rotated at training time such that the model can learn all appearances of an object. This yields good results and can scale to any problem size. It is, however, still an inefficient method because the different appearances of a single object are learned individually, which artificially inflates the complexity of any given problem. This results in slower training and, consequently, lower final performance (see Figure \ref{fig_data_aug_training_time}).

It is a long-standing conjecture in neuro-psychology that when humans try to identify an object, they mentally simulate rotations of that object with the goal of matching it to an internal representation (i.e. they perform mental rotation). \citet{shepard1971mental} were the first to formally study this phenomenon. They were able to show that the time human subjects need to determine whether pairs of 3D figures have the same shape grows linearly with the angle of rotation between the two objects. This strongly suggests that humans perform mental rotation; otherwise, the re-identification task would be completed in constant time across angles.

Inspired by this finding, we propose a third way for deep neural networks to deal with rotated inputs that we dub \emph{artificial mental rotation} (AMR). The core idea of AMR is to first find the angle of rotation of a given input and then rotate it back to its canonical appearance before further processing, thus performing an artificial version of mental rotation. This has the advantage that the underlying method of visual recognition itself does not have to be hardened against rotations, therefore all models (even trained ones) can be used in conjunction with an AMR module without any altering. 

In short, our core contributions are: (a) We introduce the concept of mental rotation to deep learning, (b) we present a simple neural network architecture that implements AMR and can be paired with all common CNNs and ViTs, (c) we extensively test the merits of AMR on ImageNet, Stanford Cars, and Oxford Pet, and conclude that it significantly outperforms data augmentation, the current state-of-the-art, (d) we present AMR results on MNIST to enable the comparison with computationally expensive alternatives, (e) we study the viability of AMR in a scenario where only parts of the test data are rotated, (f) we present comprehensive ablation studies proofing that our trained AMR modules work in practice on synthesized as well as naturally rotated data, and (g) we show the easy transferability of a trained AMR module by moving it to a downstream task (in this case semantic segmentation), significantly increasing the performance of an existing model on rotated data.

\section{Related Work}
\label{sec_relatedwork}
There are ongoing efforts to incorporate rotation invariance (or in some cases equivariance) directly into the architectures of deep neural networks, especially for CNNs. \citet{dieleman2015rotation} introduced a rotation invariant CNN system for galaxy morphology prediction that uses multiple rotated and cropped snippets of the same image as input. \citet{DBLP:conf/icml/CohenW16} presented G-CNNs which are equivariant to a larger number of symmetries such as reflections or rotations. This is achieved by lifting the network features to a desired symmetry group. \citet{romero2020group} presented group equivariant vision transformers by extending the symmetry group lifting concept to self-attention. \citet{worrall2017harmonic} introduce H-Nets which replace regular CNN filters using circular harmonics.   Alternatively, \citet{DBLP:conf/iccv/GonzalezVKT17} have proposed to rotate the filters of a CNN and then apply spatial and orientation pooling to reduce and merge the resulting features. \cite{laptev2016ti} introduce a TI-pooling, which allows to pooling of the CNN outputs for an arbitrary number of different angled versions of the same input to create an equivariant feature.
All these methods share the key drawback that their memory footprint grows linearly with the angular resolution, which severely limits their practical usability.

Data augmentation \cite{baird1992document} is very widely used to improve the robustness and generalizability of vision models \cite{simard2003best}. It can even be used to harden the model against adversarial attacks \cite{shafahi2019adversarial}. Data augmentation has also been shown to be very effective for rotated inputs \cite{quiroga2020revisiting}. Data augmentation is the current de-facto standard technique for dealing with rotated data since it is easy to use and effective. However, data augmentation is not efficient because different appearances of the same object are learned independently.

There have been previous attempts to leverage the concept of mental rotation for computer vision. \citet{ding2014mental} trained a factored gated restricted Boltzmann machine to actively transform pairs of examples to be maximally similar in a feature space. \citet{boominathan2016compensating} train a shallow neural network to classify if an image is upright. They then combine this with a Bayesian optimizer to find upright images. They use this setup to improve image retrieval robustness. In the space of 3D vision a mental rotation-based approach achieved state-of-the-art performance for rotated point cloud classification \cite{fang2020rotpredictor}.





\begin{figure*}[]
\centering
\centerline{\includegraphics[width=1\linewidth]{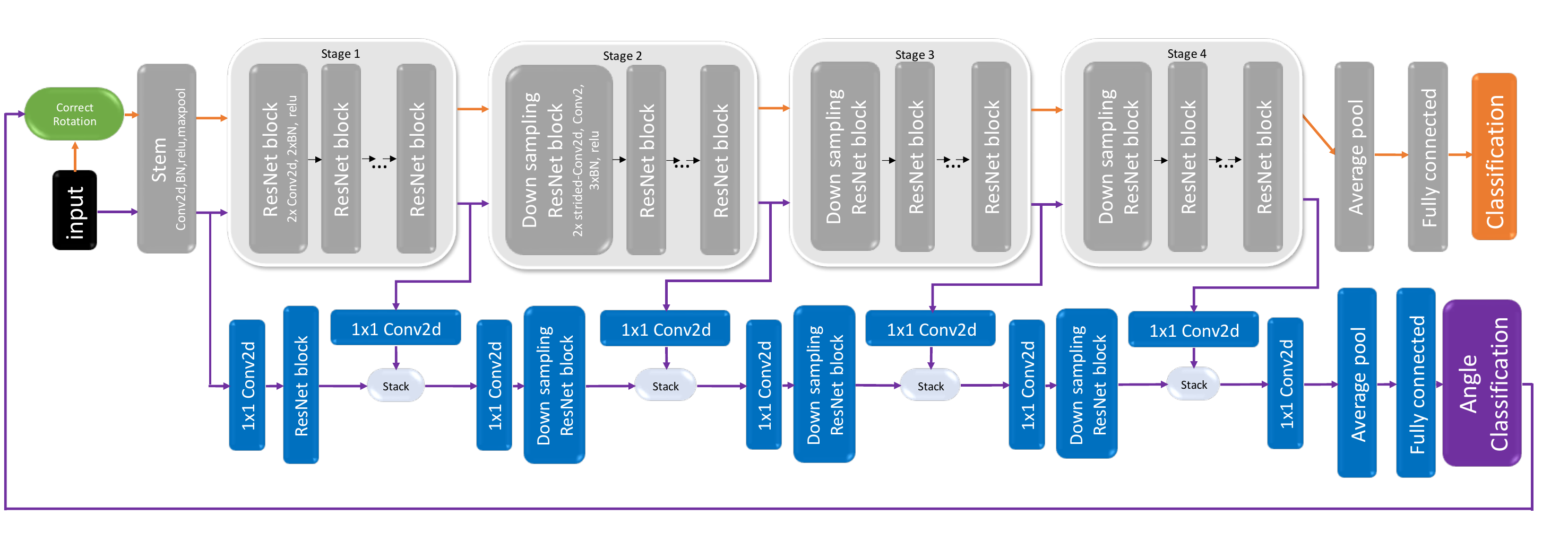}}
\caption{Architecture of our artificial mental rotation module for CNNs. The base CNN, in this case a ResNet, is shown in grey. The components of the AMR module are shown in blue. The information flow in stage 1 (angle classification) is purple while the information flow in stage 3 (image classification) is shown in orange.}
\label{fig_AMRM}
\end{figure*}

\section{Artificial Mental Rotation Module}
Our AMR approach requires three components. First, a base model (BM) is required, for which any common CNN or ViT \cite{dosovitskiy2020image} architecture can be used. There is no need to modify the BM in any way, hence the BM can generally be sourced in a fully (pre-)trained form. Additionally, it requires a rotation algorithm designed for images; here we use the method available in OpenCV \cite{opencv_library}. The last necessary component is the AMR module itself, presented in this section. 
Due to their differing designs, CNNs and ViTs use slightly varying AMR modules, described in Sections \ref{sec:AMRM_CNN} and \ref{sec:AMRM_ViT}.

\paragraph{AMR training}
While training the AMR module, the BM is frozen such that its classification performance is not disturbed. For the training, we use datasets, like ImageNet, where the objects are typically shown in an upright position. Under this constraint, we can employ self-supervised training by randomly rotating the input images and asking the AMR module to recover the angle we previously applied.

\paragraph{AMR inference}
AMR inference is performed in a three-step process: (1) The input's angle is classified by running it through the BM and the AMR module. (2) The input is rotated by the negative amount of the angle determined in step one. (3) The rotation-corrected input is processed by the BM. \\
Step (3) is identical to AMR-free inference since the BM is frozen during AMR training and there is no information flow through the AMR module during this step. Therefore AMR could also be framed as a preprocessing method by reducing it to steps (1) and (2).

\subsection{AMR module for CNNs}
\label{sec:AMRM_CNN}
Our AMR module is designed as an add-on to a given BM (see Figure \ref{fig_AMRM}), so it can repurpose the features computed by the BM and only requires a small number of additional weights. Features are copied into the AMR module at five different BM stages. In the case of ResNe(X)ts \cite{he2016deep, xie2017aggregated} (a.k.a. Highway Nets with open gates \cite{srivastava2015highway}) this happens directly after the stem and after each of the four ResNe(X)t stages. For EfficientNets \cite{tan2019efficientnet} we use the end of stages $2$, $4$, $6$, and $8$ as extraction points. When the copied features enter the AMR module they are first processed by a single $1 \times 1$ 2D convolution to compress the feature depth. For all but the first AMR module stages, these features are then stacked with the output of the previous AMR module stage followed by another $1 \times 1$ 2D convolution to half the feature depth of the stack. Only then the data is processed by a single ResNet block. After the last stage, we employ average pooling and a single fully connected layer with $360$ outputs to create the angle prediction.

\subsection{AMR module for ViTs}
\label{sec:AMRM_ViT}
The AMR module for ViTs functions very similarly to the one for CNNs. We again extract features at five different locations. For ViT-16-b these are after encoder blocks $1$, $4$, $7$, and $12$. Since there is no spatial downsampling in ViTs there is no advantage in processing the extracted features in stages. We, therefore, stack them all at once followed by a single $1 \times 1$ 2D convolution. This stack is then processed by four ViT encoder modules. Lastly, we extract the same classification token that was used in the BM and apply a fully connected layer for the angle classification.

\begin{table*}[]
\centering
\caption{ImageNet top-1 accuracies. Upright testing (up) of the upright trained base model is assumed to be the performance ceiling (\% ceil). Average (by angle) rotated accuracies (rot) are given for the upright and rotated trained base models as for AMR 33 epochs and AMR 5 epochs.}
\centerline{
\begin{tabular}{llllllllll}
 \toprule
&  \multicolumn{3}{l}{Upright Training}  &  \multicolumn{2}{l}{Rotated Training}  &  \multicolumn{2}{l}{AMR 33} &  \multicolumn{2}{l}{AMR 5} \\
Testing       & up  & rot & \% ceil & rot  & \% ceil & rot             & \% ceil & rot            & \% ceil \\
\hline
ResNet-18         & 0.695 & 0.433 & 0.62 & 0.598 & 0.86 & \textbf{0.676} & \textbf{0.97} & 0.666 & 0.96 \\
ResNet-50         & 0.768 & 0.537 & 0.70 & 0.673 & 0.88 & \textbf{0.755} & \textbf{0.98} & 0.746 & 0.97 \\
ResNet-152        & 0.779 & 0.552 & 0.71 & 0.730 & 0.94 & \textbf{0.767} & \textbf{0.98} & \textbf{0.760} & \textbf{0.98} \\
EfficientNet-b0   & 0.680 & 0.454 & 0.67 & 0.611 & 0.90 &\textbf{0.666} & \textbf{0.98} & 0.656 & 0.96 \\
EfficientNet-b2   & 0.692 & 0.467 & 0.67 & 0.612 & 0.88 & \textbf{0.678} & \textbf{0.98} & 0.669 & 0.97 \\
EfficientNet-b4   & 0.710 & 0.485 & 0.68 & 0.618 & 0.87 &\textbf{0.696} &\textbf{0.98} & 0.689 & 0.97 \\
ResNext-50-32x4d  & 0.773 & 0.551 & 0.71 & 0.686 & 0.89 & \textbf{0.761} &\textbf{0.98} & 0.754 & 0.98 \\
ResNext-101-32x8d & 0.785 & 0.571 & 0.73 & 0.728 & 0.93 & \textbf{0.772 }&\textbf{0.98}& \textbf{0.766} &\textbf{0.98} \\
\hline
ViT-16b & 0.691 & 0.459 & 0.66    & 0.503 & 0.73    &\textbf{0.669}  & \textbf{0.97}  & 0.664 & 0.96 \\
\hline
Average           & 0.730 & 0.501 & 0.69 & 0.640 & 0.87 & \textbf{0.716} & \textbf{0.98} & 0.708 & 0.97 \\
\bottomrule
\end{tabular}
}
\label{tbl_imagenet}
\end{table*}

\subsection{Motivation for add-on design}
We opted to design our AMR module as an add-on to existing base networks because we conjecture that the features that have been trained for classification will also be at least partly useful for angle detection and the AMR module can profit from the training resources that have already been invested into the base network. This design choice therefore allows for an AMR module that consists of very few layers on its own and thus can be trained very quickly. We confirm this conjecture with an ablation study (see Appendix \ref{chp_stages}).

\section{Experiments}
We aim to showcase the merits of AMR on natural images. Therefore, we test it on ImageNet (ILSVRC 2012) \cite{imagenet15russakovsky} and verify our results on Stanford Cars \cite{KrauseStarkDengFei-Fei_3DRR2013} and Oxford Pet \cite{parkhi12a}. To ensure that artificial rotations are not obvious, we mask out the corners of all images such that a centred circle remains (see Figure \ref{fig_mask} and Section \ref{sec_sanity} for an ablation study ensuring the artificial rotations are not carrying any unwanted information). For a fair comparison between upright training (without data augmentation) and training with random rotations as input data augmentation (rotated training), we train all of our base models from scratch with this masking applied. For all of our training runs, we use image normalization based on dataset statistics. No further data augmentation is applied to keep the experiments as simple as possible (except, of course, input rotation for the rotated training models). 
To obtain representative results we replicate our experiments on a variety of base models. We use three different ResNets, three EfficentNets, and two ResNeXts for a total of eight CNN architectures. On ImageNet we also employ a vision transformer in the form of ViT-16b, which is unsuited for the other smaller datasets. For each upright trained base model, we train two AMR modules: One is trained for one-third of the base model's training time (in epochs) and the other one for one-twentieth. 

\paragraph{Training details}
For all base models, we use the implementations from the torchvision \cite{torchvision2016} Python package without any modifications. To enable optimal training speed our code is based on the ffcv library \cite{leclerc2022ffcv}. All training details and links to code and trained model weights can be found in Appendix \ref{sec:repro}.

\begin{figure}[]
\centering
\includegraphics[width=0.48\linewidth]{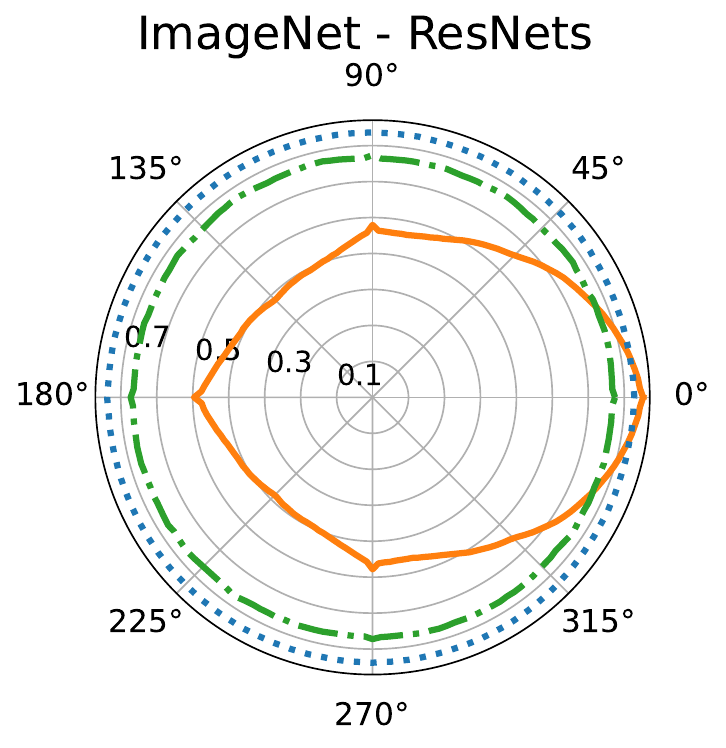}
\includegraphics[width=0.48\linewidth]{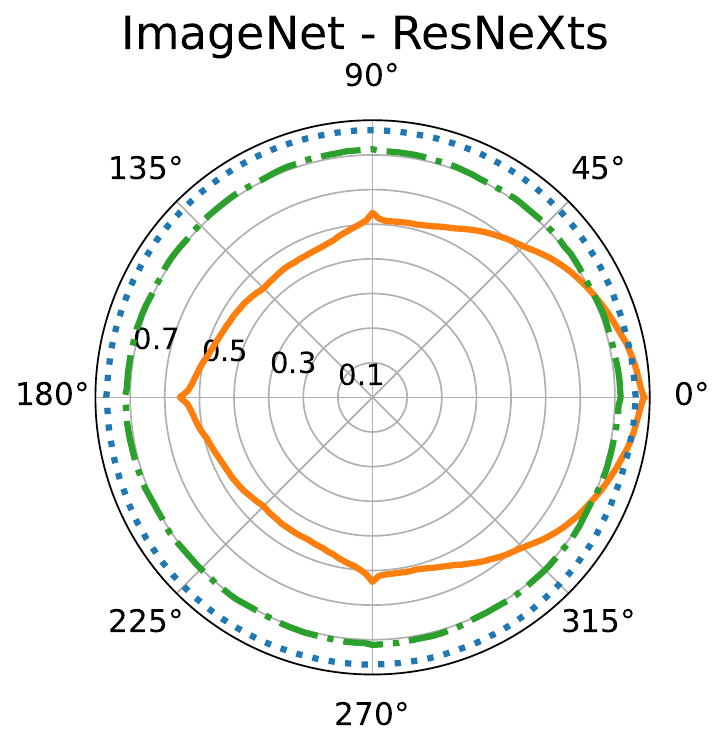}\\
\includegraphics[width=0.48\linewidth]{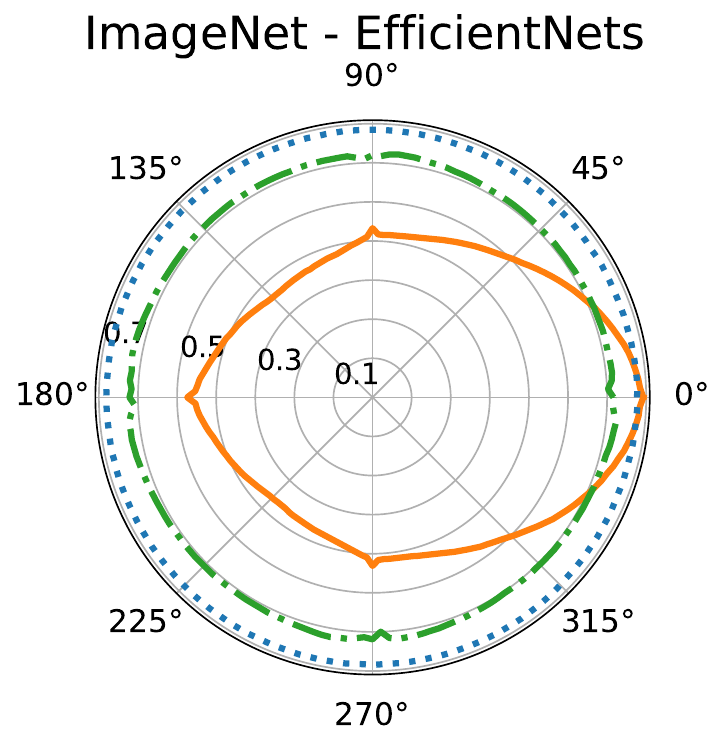}
\includegraphics[width=0.48\linewidth]{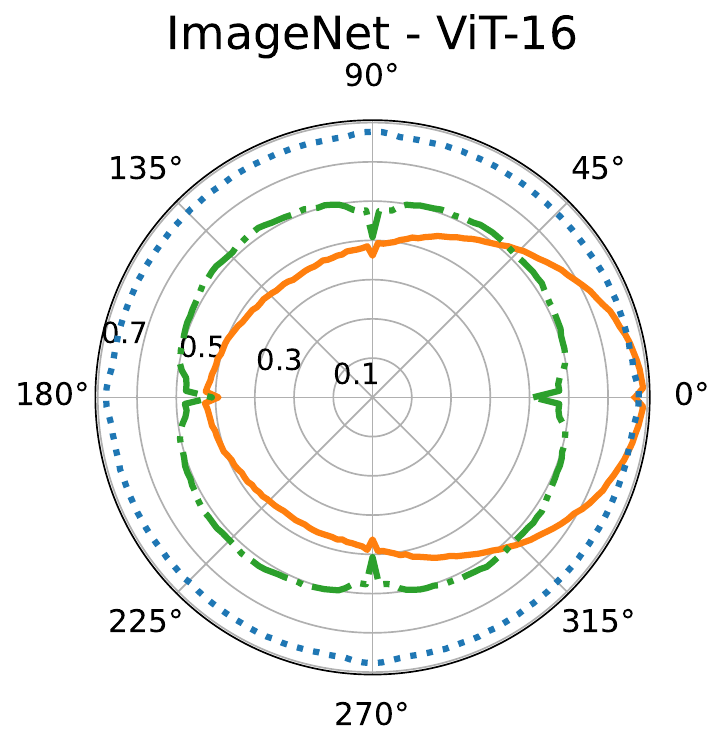}
\caption{Polar plots of ImageNet top-1 accuracies by angle, averaged across architectures (except ViT). The performances of the upright base models are shown in solid orange, the rotated training base models are shown in dash-dotted green, and AMR performance (averaged across both epoch regimes) is shown in dotted blue lines.}
\label{fig_polar_imagenet}
\end{figure}

\begin{figure}[ht!]
\centering
\includegraphics[width=0.48\linewidth]{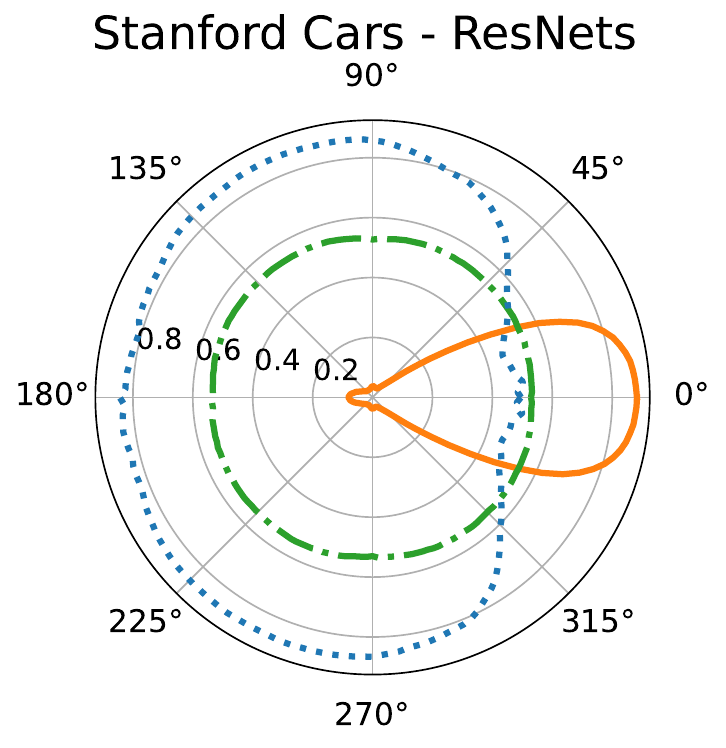}
\includegraphics[width=0.48\linewidth]{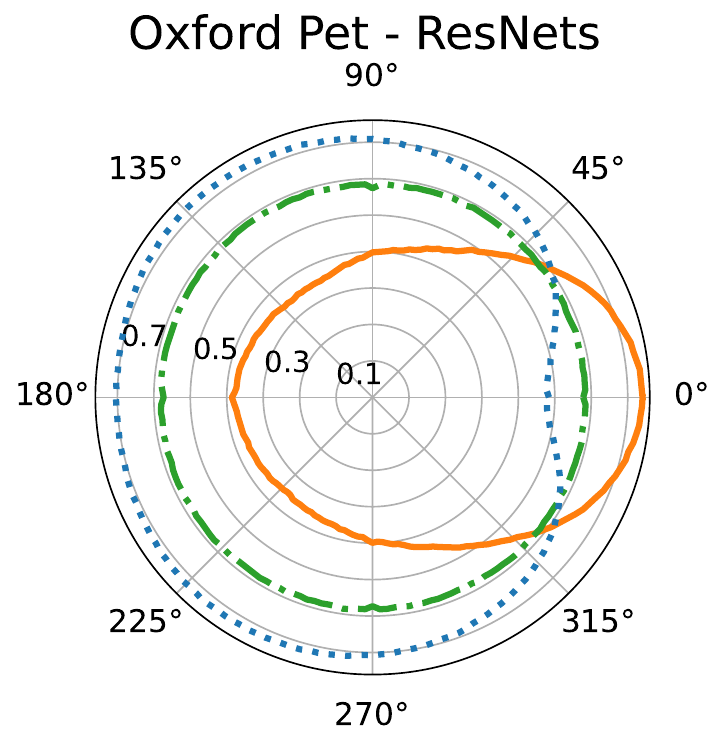} \\
\includegraphics[width=0.48\linewidth]{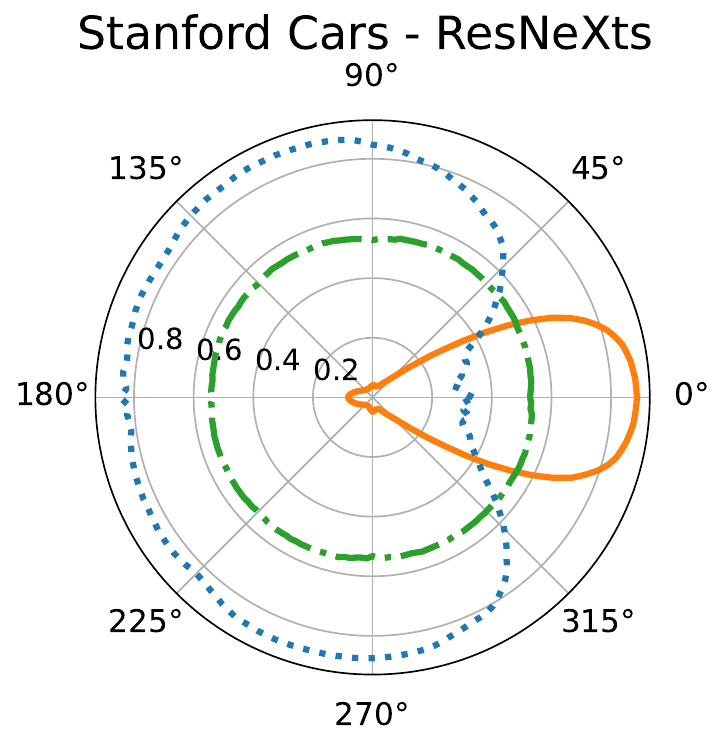}
\includegraphics[width=0.48\linewidth]{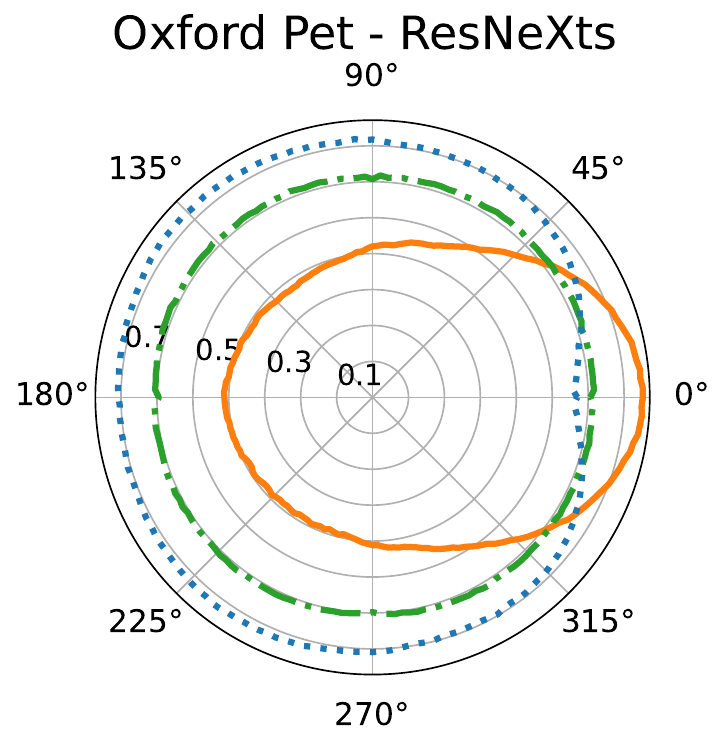} \\
\includegraphics[width=0.48\linewidth]{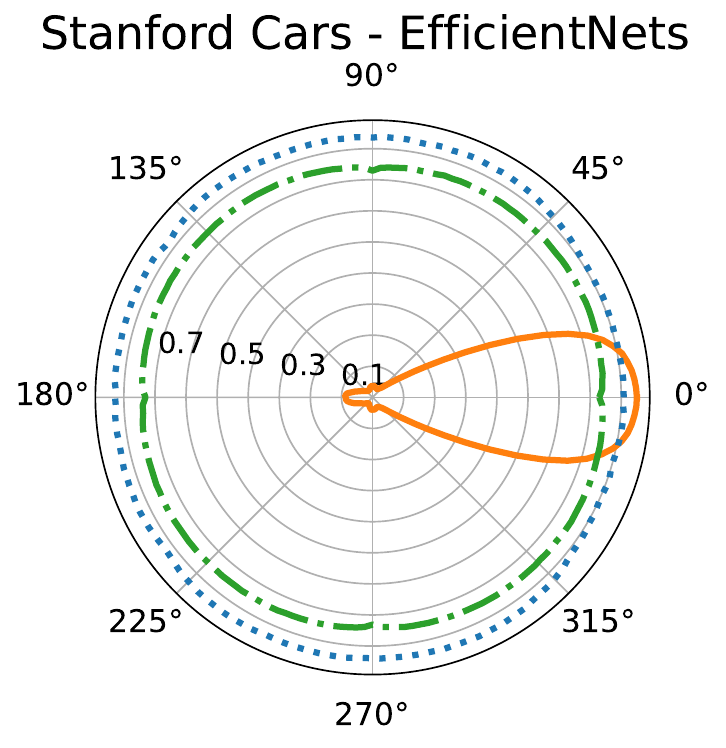} 
\includegraphics[width=0.48\linewidth]{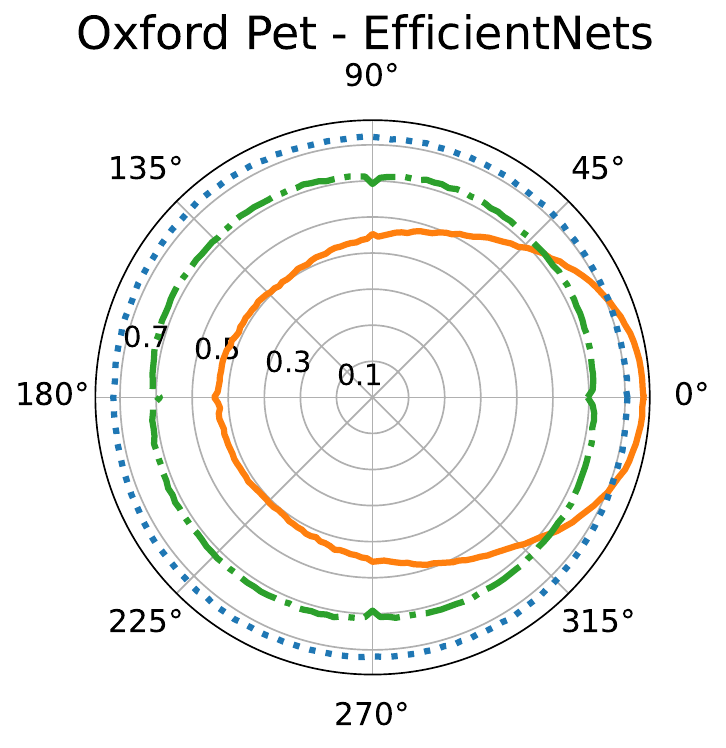}
\caption{Polar plots of Stanford Cars (left column) and Oxford Pet (right column) top-1 accuracies analogous to the ones shown above for ImageNet (see Figure \ref{fig_polar_imagenet}).}
\label{fig_stanford_cars}
\end{figure}

\paragraph{Testing}
We first evaluate the upright base models on upright data. We use these performances as the ceiling of what can be achieved on rotated data. Then we test the upright and rotated base models as well as the AMR-enhanced models for rotated performance by rotating the test set two degrees at a time and running a full evaluation for each angle. We present the resulting data visually in polar plots (see Figure \ref{fig_polar_imagenet} and Figure \ref{fig_stanford_cars}) as well as in table form (see Table \ref{tbl_imagenet} and Table \ref{tbl_stanford_cars}) by averaging across angles.

\begin{table*}[ht]
\centering
\caption{Stanford Cars  and Oxford Pet top-1 accuracies averaged across all architectures, columns are analogous to the Table \ref{tbl_imagenet} shown above for ImageNet.}
\centerline{
\begin{tabular}{llllllllll}
 \toprule
&  \multicolumn{3}{l}{Upright Training}  &  \multicolumn{2}{l}{Rotated Training}  &  \multicolumn{2}{l}{AMR 300} &  \multicolumn{2}{l}{AMR 50} \\
Testing       & up  & rot & \% ceil & rot  &\% ceil & rot& \% ceil & rot            & \% ceil \\
\hline
Stanford Cars           & 0.867 & 0.165 & 0.19 & 0.618 & 0.71 & \textbf{0.796} & \textbf{0.92} & 0.746 & 0.86 \\
Oxford Pet           & 0.741 & 0.483 & 0.65 & 0.603 & 0.81 & \textbf{0.712} & \textbf{0.96} & 0.670 & 0.90 \\
\bottomrule
\end{tabular}
}
\label{tbl_stanford_cars}
\end{table*}

\paragraph{ImageNet}
We train all of our base CNNs for $100$ epochs on ImageNet, and the vision transformer is trained for $300$ epochs, in accordance with the training recipes for the torchvision base models.  We then train two AMR modules in conjunction with each upright trained base model, one for $33$ epochs and the other for $5$. Table \ref{tbl_imagenet} contains the top-1 accuracies on rotated data for all models. Additionally, the ceiling accuracy is also reported (upright data with upright trained model). As suspected, there is a steep drop in accuracy between upright and rotated testing for the upright-trained models, both for the CNNs as well as the ViT. On average only $69$ percent of the ceiling performance ($\%$ ceil) is retained. The models which have been trained with random rotations fare much better, they achieve $87\%$ ceil. It is noteworthy that the ViT only rises from $66$ to $73$ of the ceiling performance. This makes sense since ViTs tend to be less sample efficient compared to CNNs and therefore suffer more from the increased problem complexity caused by the random rotations. AMR-33 achieves $98\%$ ceil, significantly outperforming rotated training. AMR-5 is slightly worse with $97\%$ ceil, but it shows that it is possible to obtain an AMR module that is very useful with minimal training resources.
Figure \ref{fig_polar_imagenet} contains polar plots that show the ImageNet top-1 accuracies of the different architecture families by angle. The solid orange lines show the accuracies of the upright-trained base models. We observe that the accuracies have their highest points at zero degrees rotation and then symmetrically drop off with increasing angle, reaching their lowest points at $135$ and $225$ degrees. We further observe that rotated training (green dash-dotted line) and AMR (blue dotted line) both achieve rotational invariance and exhibit performances that are independent of test time angles. Corresponding to the reported results in Table \ref{tbl_imagenet}, AMR performance is consistently better than rotated training.

\begin{table*}[ht]
\centering
\caption{Top-1 accuracies of ResNets on upright (up) and rotated (rot) ImageNet, accompanied with breakpoints (BP) that signify the share of rotated data in the test set necessary for alternative methods (rotated training, AMR) to outperform upright training. }
\begin{tabular}{lcccccccc}
\toprule
           & \multicolumn{2}{c}{Upright Training}  &  \multicolumn{3}{c}{Rotated Training}      & \multicolumn{3}{c}{AMR-33} \\
Testing    & up & rot    & up & rot & BP & up & rot & BP \\
\hline
ResNet-18  & 69.5 & 43.3    & 59.3 & 58.9 & 35.5\% & 67.1 & 67.6 & 9.0\% \\
ResNet-50  & 76.5 & 53.7    & 69.2 & 67.3 & 35.0\% & 75.1 & 75.5 & 6.1\% \\
ResNet-152 & 77.9 & 55.2    & 73.6 & 73.0 & 19.5\% & 76.2 & 76.7 & 7.4\% \\
\hline
Average & 74.6 & 50.7    & 67.4 & 66.4 & 30\% & 72.8 & 73.3 & 7.5\% \\
\bottomrule
\end{tabular}
\label{tbl_breakpoints}
\end{table*}

\paragraph{Stanford Cars and Oxford Pet} Due to Stanford Cars and Oxford Pet being smaller datasets we forgo ViTs and train the CNN models for more epochs on Stanford Cars and Oxford Pet. On Stanford Cars we train the base models for $1000$ epochs, and the corresponding AMR modules are trained for $300$ and $50$ epochs, respectively. On Oxford Pet, we train the base models for $3000$ epochs and the AMRs for $1000$ and $150$ epochs. Table \ref{tbl_stanford_cars} shows the top-1 accuracies averaged across architectures and averaged across angles where appropriate (for full table see Table \ref{tbl_stanford_cars_extended} in the Appendix) and Figure \ref{fig_stanford_cars} polar plots, analogous to the ones for ImageNet in the above paragraph. Our core findings are replicated on both datasets: Rotating the images reduces all the models' performances and AMR remains the most potent way of addressing rotations. On Stanford Cars, the performance loss caused by rotations on the upright trained model is much more severe ($19\%$ ceil) compared to ImageNet ($69\%$ ceil), with the models failing almost completely when facing rotations larger than $20$ degrees (see left column of Figure \ref{fig_stanford_cars}). This makes sense intuitively since cars are almost always upright in pictures with minimal variation, thus the models experience almost no variation during training. This is further supported by the observation that Oxford Pet, which is also a small dataset but contains animals that are naturally less static compared to cars, exhibits a milder drop off ($65\%$ ceil).
We further observe that on Stanford Cars and to a lower extent on Oxford Pet, EfficientNets perform much better than ResNe(X)ts on rotated data, both with rotated training and AMR, while all architectures perform roughly equally well on upright data. We conjecture this is because EfficientNets have been designed to be sample efficient. This could allow them to train filters useful for a wide variety of tasks (such as AMR) even on a small dataset and a relatively short training time. However, an unexpected result is that AMR paired with ResNe(x)t models showed a decline in performance when approaching $0$ degrees, while EfficientNets do not suffer from this effect. See Appendix \ref{sec_cars_pets_extended} for further investigation of this phenomenon.

\begin{table}[]
\caption{Top-1 accuracies on rotated MNIST for ResNet-18 based methods as well as related works, accompanied by ResNet-18 upright top-1 accuracy as a baseline.}
\begin{tabular}{lc}
\toprule
Method                       & Top-1 Acc. \\
\hline
ResNet-18 (upright - ceil performance) & 0.996 \\
\hline
ResNet-18                    & 0.48  \\
ResNet-18 + rotated training & 0.978 \\
ResNet18 + AMR                 & 0.981 \\
Harmonic Networks \cite{worrall2017harmonic} & 0.983 \\
Ti-pooling \cite{laptev2016ti} & 0.988 \\
G-CNNs \cite{DBLP:conf/icml/CohenW16} & 0.977 \\
RotEqNet \cite{DBLP:conf/iccv/GonzalezVKT17} & 0.989 \\ 
\bottomrule
\end{tabular}
\label{tbl_mnist}
\end{table}

\paragraph{Comparison with existing rotation equivariant methods} The focus of AMR is large datasets like ImageNet and beyond. The current literature for rotation equivariant methods is focused on computationally expensive methods that re-engineer the basic structure of the used neural networks (as mentioned in Section \ref{sec_relatedwork}). Consequently, MNIST is the benchmark dataset of choice for most of these methods. We put our work into perspective with these related works by presenting the performances of ResNet-18, ResNet-18 + rotated training and ResNet18+AMR on MNIST (see Table \ref{tbl_mnist}). The ceiling performance of ResNet18 on upright MNIST is almost one, which is to be expected. Similar to the larger datasets above is the performance of AMR far superior to rotated training. Most importantly, the performance of ResNet18+AMR is comparable to the ones of the related works which are much narrower in scope. This shows that AMR not only exhibits the best performance on large datasets but also achieves state-of-the-art performance on a small dataset amongst highly specialized methods.

\paragraph{AMR usefulness given the prevalence of rotated data} In an applied scenario, it is not always realistic that all inputs are presented at a random angle. We therefore investigate the usefulness of AMR when the test data consists of a combination of upright (up) and rotated (rot) images. To this end, we compute top-1 test errors on ImageNet of the ResNet family models on rotated and upright inputs separately. We repeat this process for upright training, rotated training and AMR-33 (see Table \ref{tbl_breakpoints}). We then linearly combine up and rot performances to obtain the final performances for mixed datasets consisting of both upright and rotated data. We increase the percentage of rotated data in the test mix until alternative methods (rotated training, AMR-33) start outperforming the default of upright training. We call percentages of parity between methods breakpoints (BP). Unsurprisingly, the BPs for rotated training ($30\%$ on average) are much higher than the ones of AMR-33 ($7.5\%$). The key finding here is that BPs for AMR-33 are all below $10\%$ which shows that only a small portion of the test set needs to be non-upright for AMR to be a worthwhile choice.

\begin{figure*}[ht]
\centering
\includegraphics[width=.8\linewidth]{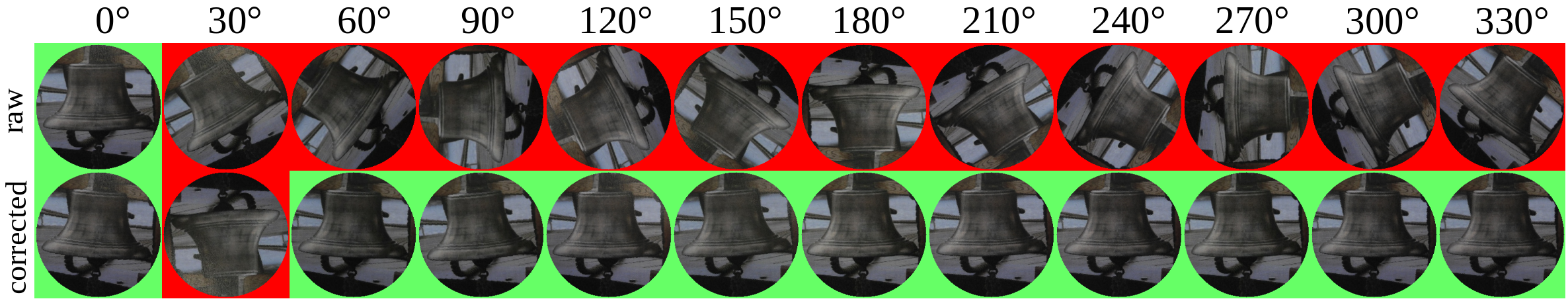}
\includegraphics[width=.8\linewidth]{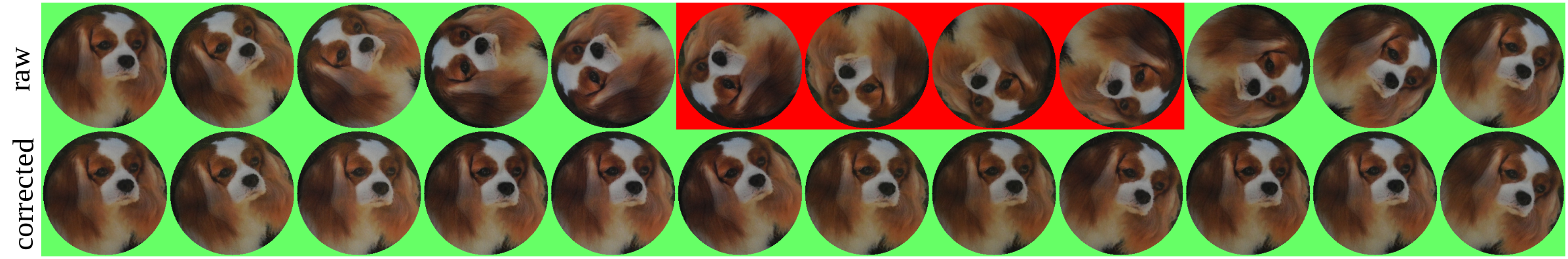}
\caption{Photographs of two printed ImageNet validation samples taken at $12$ different angles. Both samples are shown in their original state (raw) and after the mental rotation step (corrected). The color of the masked-out region indicates if the corresponding image has been correctly classified. The mental rotation and classification steps have been performed by ResNet50 + AMR33.}
\label{fig_sanity}
\end{figure*}

\section{Validity of self-supervised training built on artificial rotation }
\label{sec_sanity}
Self-supervised learning based on artificial data modifications always warrants great caution. It is often unclear if the model learns to solve the desired task or if it simply learns to find unintended shortcuts in the self-supervision procedure. In our case, we use a digital rotation algorithm on our input images. While none are visible to the human eye, algorithm-specific artefacts are introduced to the rotated images. This raises the question if the AMR module learns to classify the correct rotation angle based on unwanted traces of the rotation algorithm rather than by understanding the contents of the image. To ensure this is not the case, we perform the following ablation study: We print out seven images sourced from different classes from the ImageNet validation set. We then take photos of each of those printouts at twelve different in-plane rotations by physically rotating the print in $30$-degree intervals. This way we naturally introduce rotation and can guarantee the absence of any rotation algorithm artifacts that the model could have learned to use. The images were printed using a Konica Minolta bizhub 450i on maximum resolution with guidelines to enable accurate angular distances (see Figure \ref{fig_printout} in the Appendix). The photos were then taken by hand using a Nikon Coolpix P7000 digital camera.
Figure \ref{fig_sanity} shows all twelve re-digitized photos for two cases (raw). The color-coded background indicates if that photo was correctly classified by a standard trained ResNet50 base model (green denotes correct, red an error). We observe a similar effect as with Stanford Cars: Like a car, bells have a very clearly defined upright position. The bell, therefore, is only classified correctly when it is upright. Dogs on the other hand are very variable in appearance (e.g. head turned, laying down etc), thus the dog is only misclassified when it is completely upside down at rotations between 150 and 240 degrees. The second rows (corrected) show the outcome of applying Resnet50 + AMR33 to the above photos. The AMR module is able to correct the orientation of all but one photo. We can therefore conclude that it learned to classify the angles by understanding the image contents rather than relying on artifacts introduced by the self-supervision process. We further observe that the rotation correction is much more precise in the bell case than for the dog. This ties in with our assumption that the network's filters are much more precisely tuned to a sharp upright position for the bell compared to the dog.
Across all $84$ photos, the standard ResNet50 achieves a top-1 classification accuracy of $0.57$. ResNet50 + AMR33, on the other hand, achieves a top-1 accuracy of $0.96$, showing that the AMR module properly works on all printed images.


\begin{figure*}[]
\centering
\includegraphics[width=0.75\linewidth]{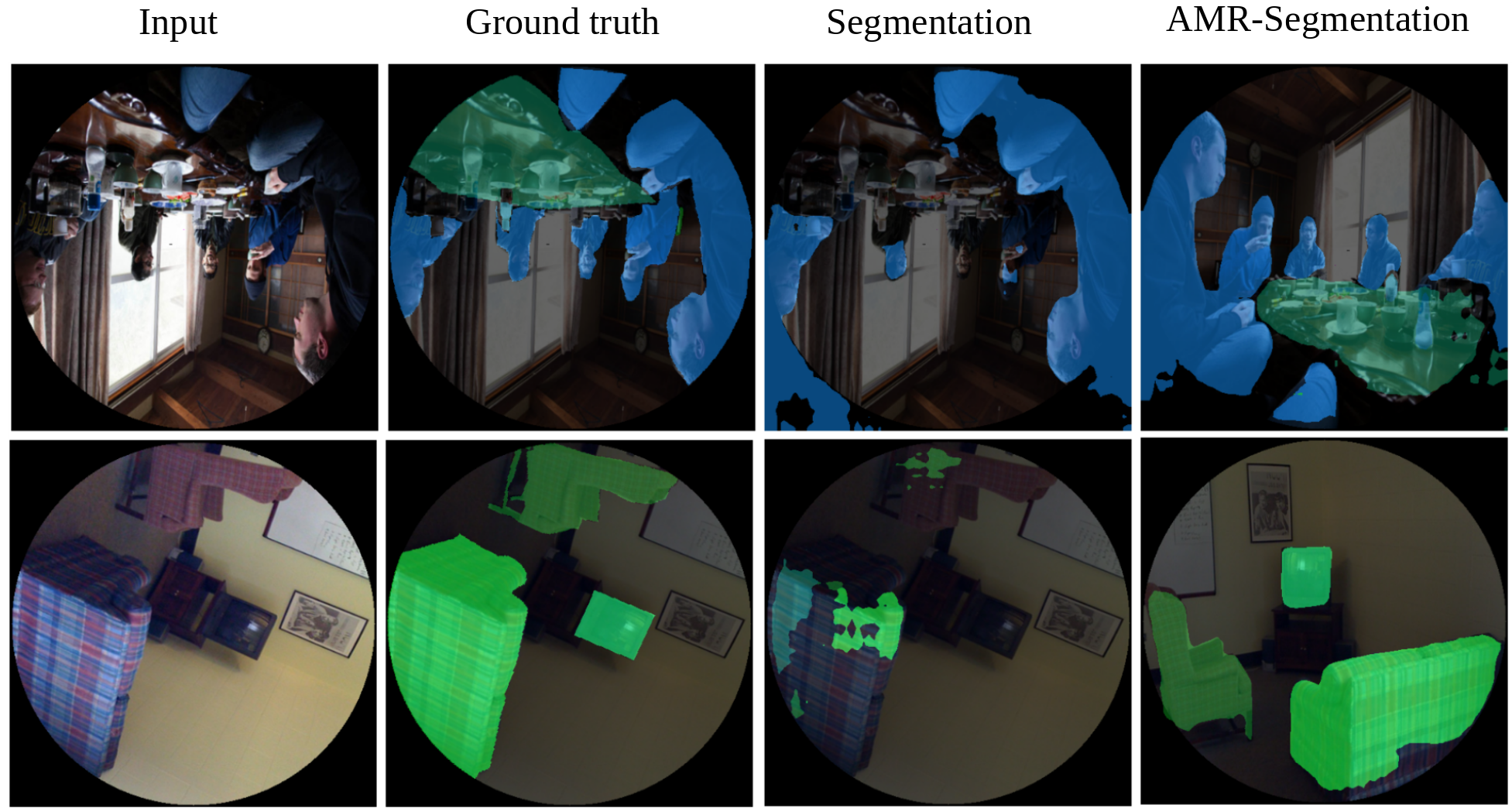}
\caption{Two rotated examples (rows) from the CoCo validation set with their corresponding ground truth, segmentation output, and AMR-corrected segmentation output (columns). The segmentation on the rotated image exhibits bad performance in both samples. In row one, the persons are segmented fairly well but the table is missing fully, in row two the segmentation fails almost completely. The column AMR-segmentation shows the output for the images that have been un-rotated using AMR. The AMR module identified the correct rotation angle in both cases, which lead to significantly improved segmentation performance (again in both cases).}
\label{fig_semseg}
\end{figure*}
\hspace{1cm}

\section{Application to a Novel Downstream Task: Semantic Segmentation}
Since they use the same neural network building blocks, the assumption that models for other vision tasks like object detection or semantic segmentation also struggle with rotated inputs suggests itself. In this section, we test this hypothesis and demonstrate how a trained AMR module can be used to easily improve the rotational stability of models for other tasks than classification. Here we choose semantic segmentation as an example. As the base model, we use a fully convolutional ResNet50 and source the matching pre-trained weights named 'FCN\_ResNet50\_Weights.COCO\_WITH\_VOC\_LABELS\_V1' from torchvision. These weights have been trained on MSCoCo-Stuff \cite{lin2014microsoft}, with a reduced class set only containing the classes that are also available in PascalVOC \cite{pascal-voc-2012}. We again mask the corners of all images. On this data, the pre-trained model achieves a mean intersection over union (IoU) of $57.6$. We then randomly rotate the images which causes the mean IoU to drop to $32.7$. This confirms that not only object classification models but also semantic segmentation and likely most other vision models perform far worse when confronted with rotated inputs. We now take our ResNet50 + AMR33 which has been trained on ImageNet and use it to perform AMR steps (1) and (2) on CoCo without any additional retraining or modification. The angle-corrected inputs are then fed back into the base semantic segmentation model. This approach yields an IoU of $55.2$, showing that AMR also works for semantic segmentation and that a trained AMR module can be easily transferred between similar datasets. Figure \ref{fig_semseg} shows two  examples from the CoCo validation set with their corresponding ground truth, segmentation output, and AMR-corrected segmentation output visually confirming our findings. For both examples, the AMR module correctly inverted the angle of the image, which (again in both cases) significantly improves the segmentation performance.

\section{Limitations and Future Work}
\label{sec_limit}
A key drawback of AMR is that two forward passes are necessary. This is part of the core design and cannot be changed. It is mitigated partially by the fact that a smaller model can be chosen in conjunction with AMR and still outperform a large model trained with rotational data augmentation due to the inefficiency of that approach resulting in a less costly forward pass even at test time. 
With applicability in mind, we opted to focus on 2D in-plane rotations of whole images featuring one dominant object.  Our work is therefore not suited for cases where multiple objects are individually rotated. This scenario could be addressed by combining a region-proposal based method such as Faster-RCNN \cite{girshick2015fast} with AMR at proposal level. In the real world, 3D objects are rotated on two axes, which can lead to much more drastic changes in appearance. Extending AMR to 3D objects would be a very promising, most natural extension of this work. 
An exciting future application for AMR models would be reducing the rotational variability of an existing dataset (e.g. ImageNet, by making all appearing objects upright). This would further disentangle the training of upright appearances from rotations which would likely lead to improved training efficiency of base models.

\section{Conclusions}
We have presented AMR, a neuropsychology-inspired approach for handling rotated data in vision systems, novel to deep learning. We have shown that AMR consistently outperforms the current standard of input data augmentation across different deep architectures and datasets. We have shown the viability of AMR in realistic cases where the data is a mixture of upright and rotated inputs. We further presented a sanity check which confirms that our self-supervised learning setup learns to identify rotations by the content of the images and not by any artifacts introduced by the training procedure. Lastly, we have shown how a trained AMR module can easily be transferred to another model built for a different task (in our case semantic segmentation) to greatly improve its rotational stability.

\section*{Acknowledgements}
This work has been financially supported by grants 34301.1 IP-ICT “RealScore” (Innosuisse) and ERC Advanced Grant AlgoRNN nr. 742870. Thanks go to Mohammadreza Amirian and Philipp Denzel for insightful discussions and Jasmina Bogojeska for the idea of the sanity check.

\balance
\bibliography{mrotation}
\bibliographystyle{icml2023}

\newpage
\appendix
\onecolumn
\section{Extended results on Stanford Cars and Oxford Pet}
\label{sec_cars_pets_extended}
\begin{figure}[]
\centering

\includegraphics[width=0.30\linewidth]{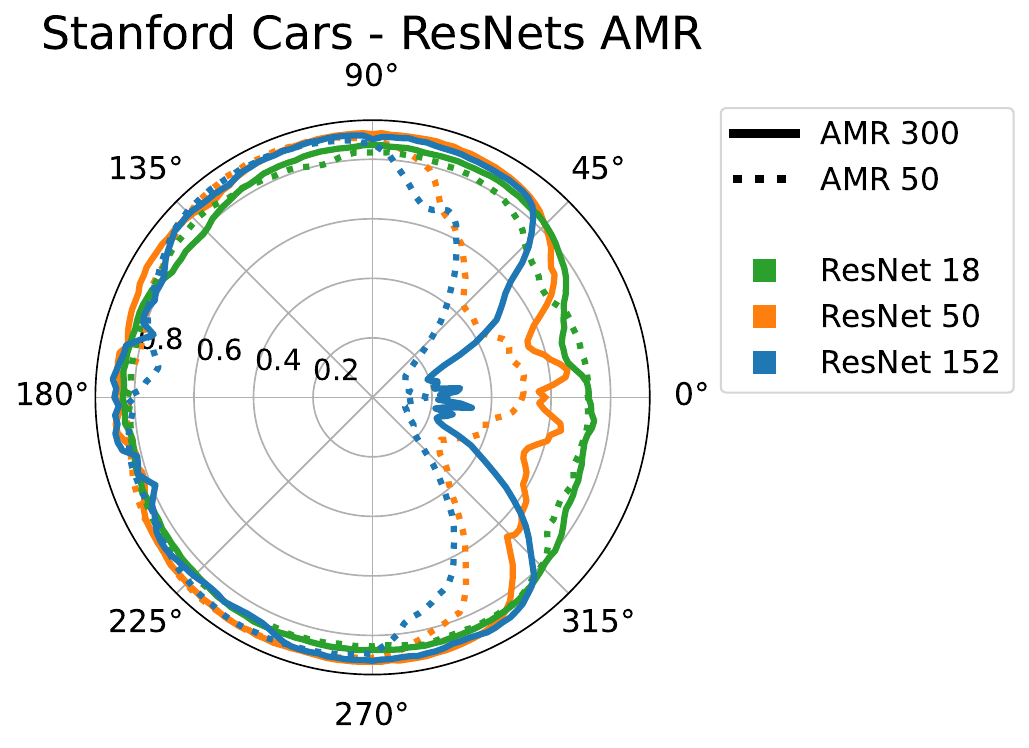}
\includegraphics[width=0.30\linewidth]{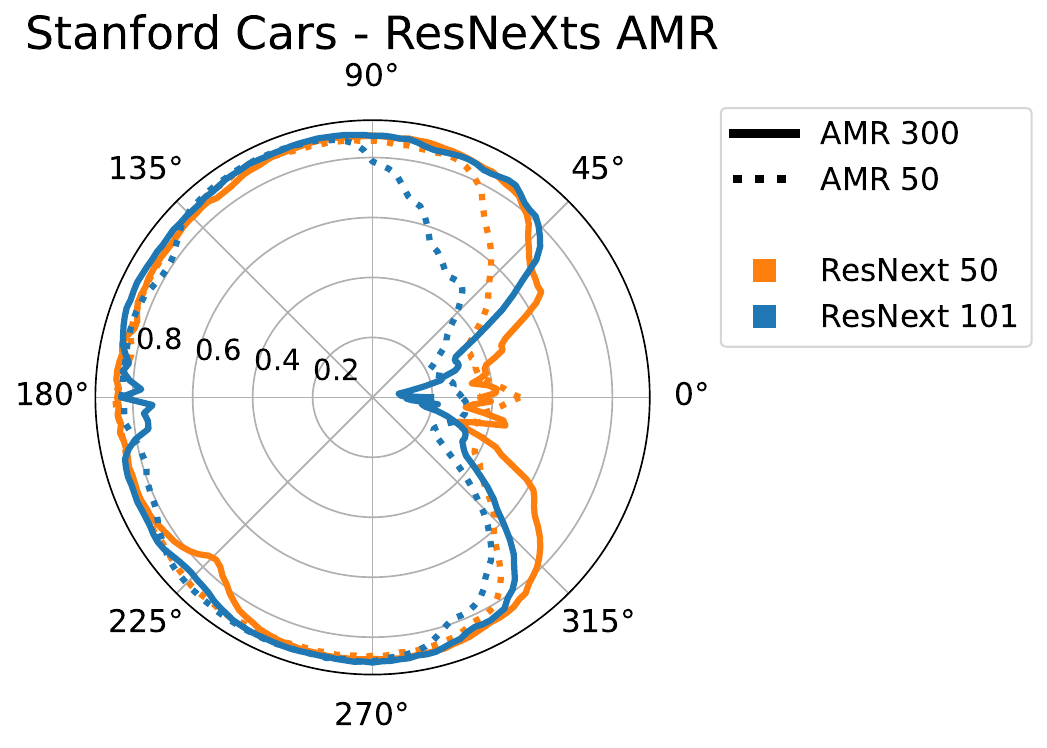}
\caption{Polar plots of Stanford Cars top-1 AMR accuracies, shown individually by training duration (dashed vs dotted) and base model architecture (color coded) instead of averaged.}
\label{fig_stanford_cars_amr}
\end{figure}

We investigate the phenomenon that for Stanford Cars (and to a lesser extent) Oxford Pet the AMR performance decreases when inputs are close to upright. We create polar plots of the Stanford Cars top-1 accuracies where AMR training duration and base architecture are shown individually (see Figure \ref{fig_stanford_cars_amr}). We observe that longer training of the AMR module helps to some extent. Using a small, data-efficient architecture such as ResNet18 on the other hand almost completely removes this effect. It shows that AMR module training is much more reliant on a sample-efficient, appropriately sized (with respect to the dataset) base model compared to standard classification training where large models tend to perform best. We, therefore, suspect that models which are oversized for a simple dataset such as Stanford Cars can learn very task-specific filters in early layers that only retain information used for classification. This in turn causes the angle classifier to fail when an upright image is used for which these filters have been optimized and very minimal information is retained in the features extracted from the base model.


\begin{table*}[]
\centering
\caption{Stanford Cars (top) and Oxford Pet (bottom) top-1 accuracies for all architectures analogous to the one shown above for ImageNet (see Table \ref{tbl_imagenet}).}
\centerline{
\begin{tabular}{llllllllll}
 \multicolumn{10}{c}{Stanford Cars} \\
 \toprule
&  \multicolumn{3}{l}{Upright Training}  &  \multicolumn{2}{l}{Rotated Training}  &  \multicolumn{2}{l}{AMR 300} &  \multicolumn{2}{l}{AMR 50} \\
Testing       & up  & rot & \% ceil & rot  &\% ceil & rot& \% ceil & rot            & \% ceil \\
\hline
ResNet-18         & 0.854 & 0.163 & 0.19 & 0.412 & 0.48 &\textbf{0.812}& \textbf{0.95} & 0.800 & 0.94 \\
ResNet-50         & 0.892 & 0.182 & 0.20 & 0.535 & 0.60 &\textbf{0.807} & \textbf{0.90} & 0.705 & 0.79 \\
ResNet-152        & 0.886 & 0.169 & 0.19 & 0.671 & 0.76 & \textbf{0.737} & \textbf{0.83} & 0.615 & 0.69 \\
EfficientNet-b0   & 0.825 & 0.140 & 0.17 & 0.687 & 0.83 & \textbf{0.820} & \textbf{0.99} & 0.785 & 0.95 \\
EfficientNet-b2   & 0.831 & 0.138 & 0.17 & 0.770 & 0.93 &\textbf{0.823} & \textbf{0.99} & 0.808 & 0.97 \\
EfficientNet-b4   & 0.885 & 0.163 & 0.18 & 0.786 & 0.89 & \textbf{0.877} &\textbf{0.99}& 0.860 & 0.97 \\
ResNext-50-32x4d  & 0.877 & 0.185 & 0.21 & 0.507 & 0.58 & \textbf{0.763} & \textbf{0.87} & 0.728 & 0.83 \\
ResNext-101-32x8d & 0.885 & 0.179 & 0.20 & 0.577 & 0.65 & \textbf{0.725} &\textbf{0.82} & 0.666 & 0.75 \\
\hline
Average           & 0.867 & 0.165 & 0.19 & 0.618 & 0.71 & \textbf{0.796} & \textbf{0.92} & 0.746 & 0.86 \\
\bottomrule \\
\multicolumn{10}{c}{Oxford Pet} \\
 \toprule
&  \multicolumn{3}{l}{Upright Training}  &  \multicolumn{2}{l}{Rotated Training}  &  \multicolumn{2}{l}{AMR 1000} &  \multicolumn{2}{l}{AMR 150} \\
Testing       & up  & rot & \% ceil & rot  & \% ceil & rot             & \% ceil & rot  & \% ceil \\
\hline
ResNet-18         & 0.700 & 0.442 & 0.63 & 0.579 & 0.83 &\textbf{0.665} & \textbf{0.95} & 0.624 & 0.89 \\
ResNet-50         & 0.750 & 0.477 & 0.64 & 0.590 & 0.79 & \textbf{0.712} & \textbf{0.95} & 0.660 & 0.88 \\
ResNet-152        & 0.752 & 0.460 & 0.61 & 0.579 & 0.77 & \textbf{0.697} & \textbf{0.93} & 0.633 & 0.84 \\
EfficientNet-b0   & 0.740 & 0.496 & 0.67 & 0.630 & 0.85 & \textbf{0.727}& \textbf{0.98} & 0.699 & 0.94 \\
EfficientNet-b2   & 0.763 & 0.518 & 0.68 & 0.617 & 0.81 & \textbf{0.753} & \textbf{0.99} & 0.720 & 0.94 \\
EfficientNet-b4   & 0.735 & 0.515 & 0.70 & 0.605 & 0.82 & \textbf{0.719} & \textbf{0.98}& 0.686 & 0.93 \\
ResNext-50-32x4d  & 0.743 & 0.489 & 0.66 & 0.579 & 0.78 & \textbf{0.711} & \textbf{0.96} & 0.679 & 0.91 \\
ResNext-101-32x8d & 0.748 & 0.469 & 0.63 & 0.645 & 0.86 & \textbf{0.712}& \textbf{0.95} & 0.656 & 0.88 \\
\hline
Average           & 0.741 & 0.483 & 0.65 & 0.603 & 0.81 & \textbf{0.712} & \textbf{0.96} & 0.670 & 0.90 \\
\bottomrule
\end{tabular}
}
\label{tbl_stanford_cars_extended}
\end{table*}

\section{Are the base network features useful for rotation estimation?}
\label{chp_stages}
\begin{figure}[]
\centering
\includegraphics[width=0.5\linewidth]{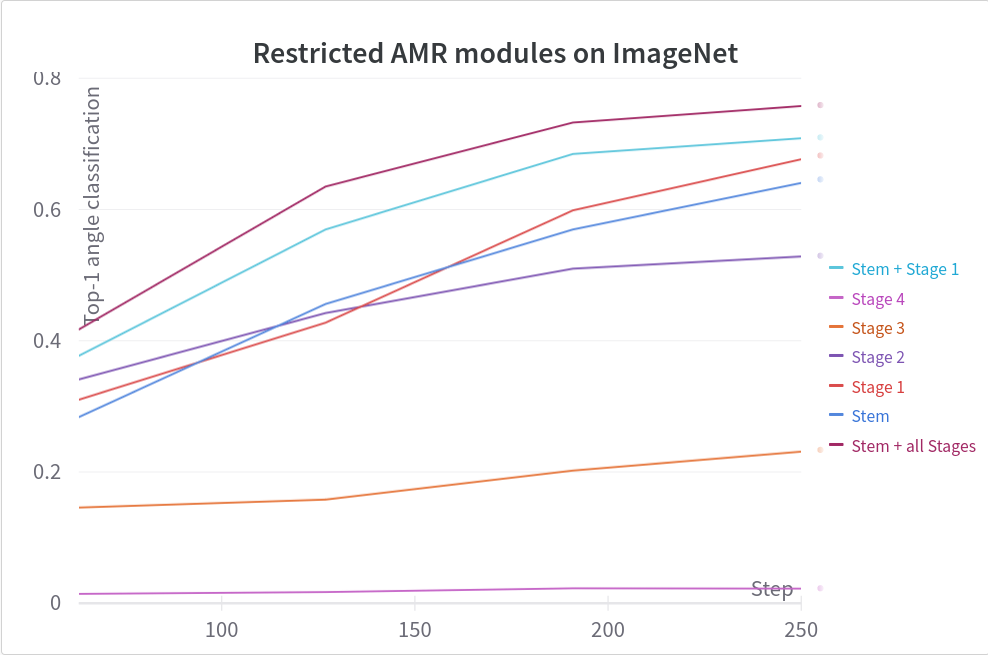}
\caption{Top-1 angle accuracies for ResNet-18+AMR with the information flow from the base network restricted. Legend indicates which channels (see Figure 2 in the main manuscript)  from the base network are open.}
\label{fig_restricted}
\end{figure}

We opt to use the features generated by a base classification network as the input to our AMR module instead of using a distinct specialized network for angle prediction. This allows us to design an AMR module with a very low number of weights such that it is quickly trained. The intuition behind this choice is that the features which are trained for classification also carry valuable information for angle prediction. This should be especially true for the early layers which tend to consist more of lower-level, class-agnostic, features. In this section, we present an ablation study exploring the validity of this assumption. We train AMR modules attached to a ResNet-18 with restricted information flow from the base model to the AMR. First, we train a module and only open the connection from the Stem to the AMR module (see Figure \ref{fig_AMRM}). We repeat this for each of the four other connections (Stage 1 - Stage 4). Figure \ref{fig_restricted} shows the evolution of the top-1 angle classification errors over the training time. The best single source of information is the output of Stage 1. A close second is the module that is only attached to the stem (making it essentially a separate network). As expected are the outputs of the later stages far less informative. We train an additional module with input from Stage 1 plus Stem, it outperforms the other modules which are attached to only one of those quite significantly. This shows that the channels carry some complementary information. Finally, the module attached to the base network at all five connections clearly performs the best. Therefore we can conclude that the features of the base network are helpful for angle prediction and feeding the output of all base network stages into the AMR module is the best design choice.

\section{Reproducibility notice}
\label{sec:repro}
This section provides all the necessary information to reproduce every result presented in this work. Our experiments were run on machines with 4xTesla V100-SXM2-32GB or 4xNVIDIA Tesla T4 16GB. When running on a single GPU we recommend the linear learning rate scaling rule \cite{DBLP:journals/corr/GoyalDGNWKTJH17}.

\textbf{Data}\\
ImageNet, Stanford Cars and Oxford Pet are public datasets and can be sourced from their original authors. Our photographed-at-an-angle versions of ImageNet images can be downloaded here: \url{https://github.com/tuggeluk/ffcv-imagenet/tree/rotation_module/sanity_check_data}.

\textbf{Code}\\
Our code is publicly available on GitHub at \url{https://github.com/tuggeluk/ffcv-imagenet/tree/rotation_module}.

\textbf{Hyperparameters}\\
We logged hyperparameters as well as relevant metrics of each run to \url{wandb.ai}. The easiest and safest way to exactly replicate a run is to first check out the git state and then use the run command given in the overview tab of each run. The logs of each training and evaluation run can be found under the URLs shown in Table \ref{tbl:hp}.

\begin{table}[]
\centering
\caption{Links to \url{wandb.ai} projects containing hyperparameters and relevant metrics of all training and evaluation runs created for this work.}
\label{tbl:hp}
\resizebox{\columnwidth}{!}{%
\begin{tabular}{ll}
\midrule
\textbf{ImageNet   CNN} &                                                                                                                                           \\
\textit{Training}       &                                                                                                                                           \\
Base models             & \url{https://wandb.ai/tuggeluk/train\_base\_models}                                                                                             \\
AMR modules             & \url{https://wandb.ai/tuggeluk/train\_angleclass\_no\_lossshape}                                                                                \\
\textit{Evaluation}     &                                                                                                                                           \\
Base models             & \url{https://wandb.ai/tuggeluk/evaluate\_final\_base\_models\_highres}                                                                          \\
BM + AMR                & \url{https://wandb.ai/tuggeluk/evaluate\_final\_angle\_class\_highres},  \\
                        &  \url{https://wandb.ai/tuggeluk/evaluate\_final\_angle\_class\_highres\_5ep} \\
\midrule
\textbf{ImageNet ViT}   &                                                                                                                                           \\
\textit{Training}       &                                                                                                                                           \\
Base models             & \url{https://wandb.ai/tuggeluk/base\%20ViT}                                                                                                     \\
AMR modules             & \url{https://wandb.ai/tuggeluk/train\_angleclass\_no\_lossshapeViT}                                                                             \\
\textit{Evaluation}     &                                                                                                                                           \\
Base models             & \url{https://wandb.ai/tuggeluk/evaluate\_final\_base\_models\_highres\_ViT}                                                                     \\
BM + AMR                & \url{https://wandb.ai/tuggeluk/evaluate\_angle\_class\_ViT}                                                                                     \\
\midrule
\textbf{Stanford Cars}  &                                                                                                                                           \\
\textit{Training}       &                                                                                                                                           \\
Base models             & \url{https://wandb.ai/tuggeluk/train\_base\_models\_stanfordcars}                                                                               \\
AMR modules             & \url{https://wandb.ai/tuggeluk/train\_angleclass\_no\_lossshape\_StanfordCars}                                                                  \\
\textit{Evaluation}     &                                                                                                                                           \\
Base models             & \url{https://wandb.ai/tuggeluk/evaluate\_final\_base\_models\_highres\_StanfordCars}                                                            \\
BM + AMR                & \url{https://wandb.ai/tuggeluk/test\_angleclass\_stanfordcars}                                                                                  \\
\midrule
\textbf{Oxford Pet}     &                                                                                                                                           \\
\textit{Training}       &                                                                                                                                           \\
Base models             & \url{https://wandb.ai/tuggeluk/train\_base\_models\_oxfordpet}                                                                                  \\
AMR modules             & \url{https://wandb.ai/tuggeluk/train\_angleclass\_no\_lossshape\_OxfordPet}                                                                     \\
\textit{Evaluation}     &                                                                                                                                           \\
Base models             & \url{https://wandb.ai/tuggeluk/evaluate\_final\_base\_models\_highres\_OxfordPet}                                                               \\
BM + AMR                & \url{https://wandb.ai/tuggeluk/test\_angleclass\_oxfordpets}                                                                                   \\
\midrule
\textbf{MNIST}     &                                                                                                                                           \\
\textit{Training}       &                                                                                                                                           \\
Base models             & \url{https://wandb.ai/tuggeluk/train_base_models_MNIST}                                                                                  \\
AMR modules             & \url{https://wandb.ai/tuggeluk/train_angleclass_no_lossshape_MNIST}                                                                     \\
\bottomrule

\end{tabular}
}
\end{table}

\textbf{Model weights}\\
The model weights for each trained model we use here can be obtained using the following link \url{https://github.com/tuggeluk/ffcv-imagenet/tree/rotation_module/downloads.md}.

\section{Re-digitized Images}
\begin{figure}[]
\centering
\includegraphics[width=0.75\linewidth]{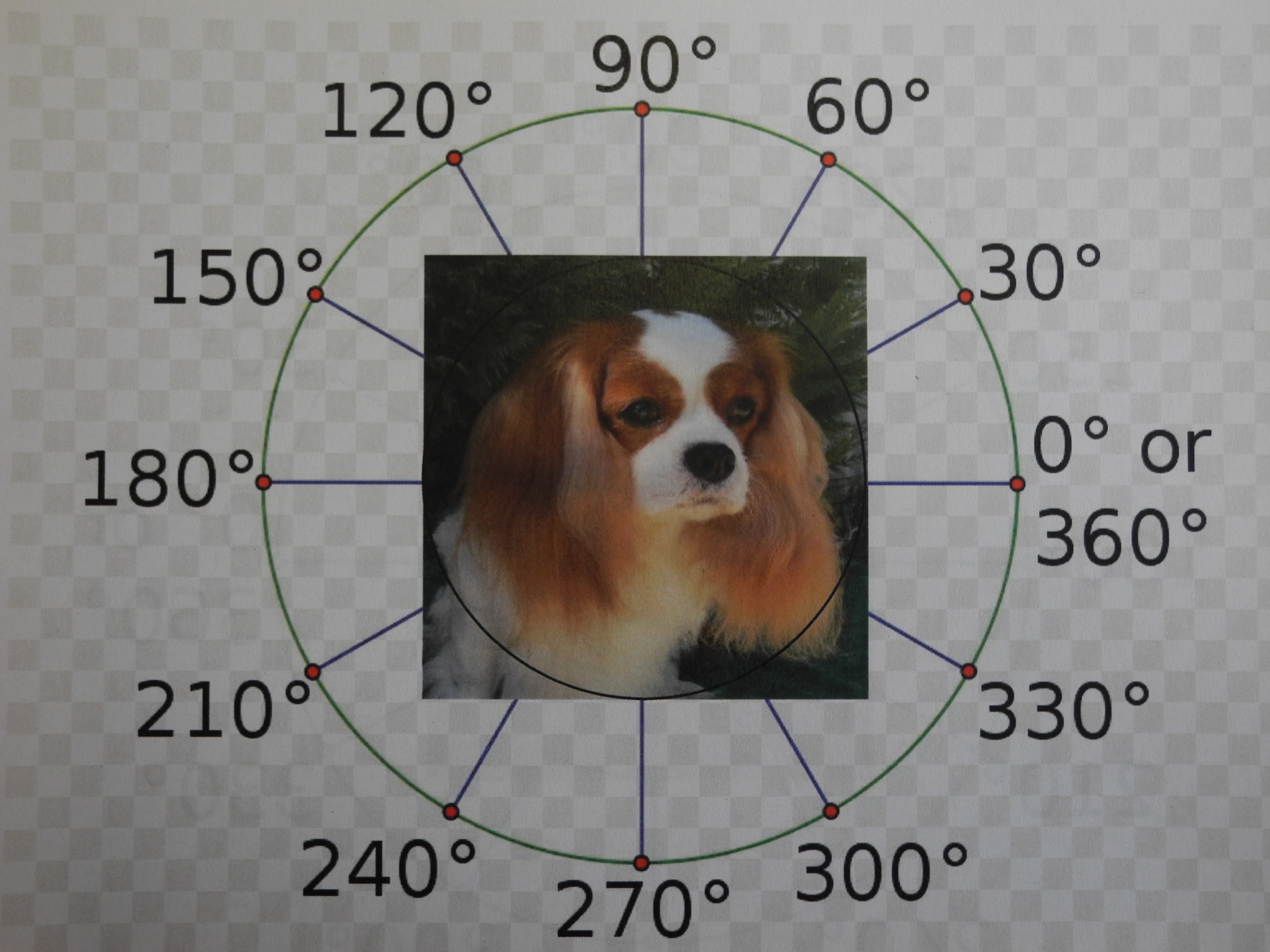}
\caption{Example printout featuring helper lines to facilitate photographs at exact angles.}
\label{fig_printout}
\end{figure}

\begin{figure}[]
\includegraphics[width=\linewidth]{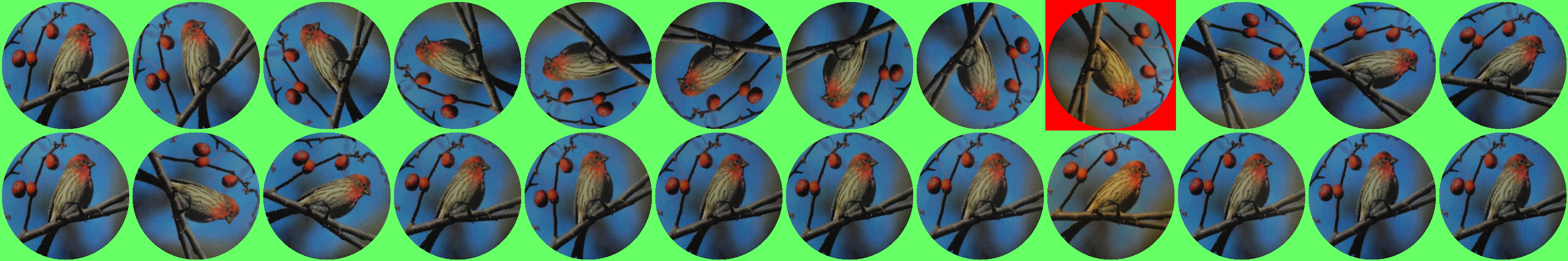}
\vspace{0.01cm}\\
\includegraphics[width=\linewidth]{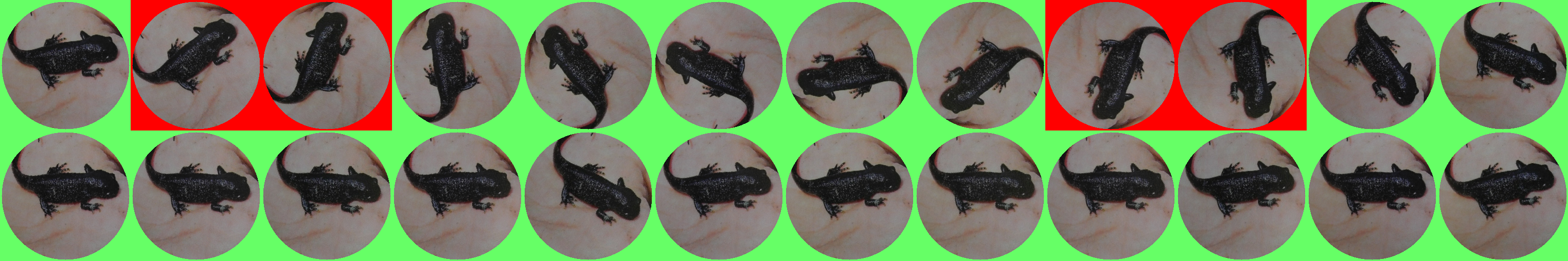}
\vspace{0.1cm}\\
\includegraphics[width=\linewidth]{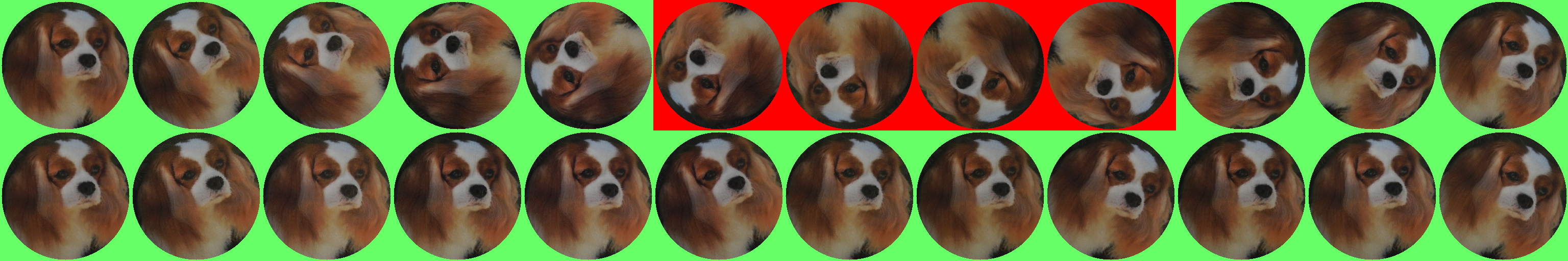}
\vspace{0.1cm}\\
\includegraphics[width=\linewidth]{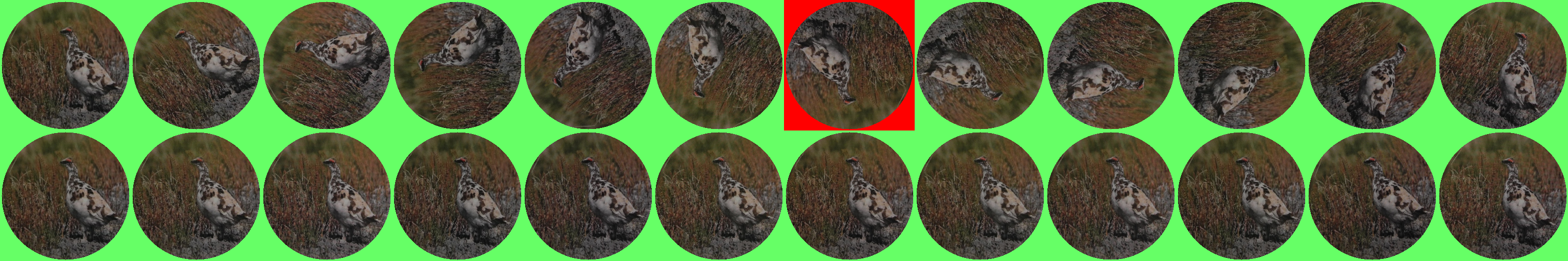}
\vspace{0.1cm}\\
\includegraphics[width=\linewidth]{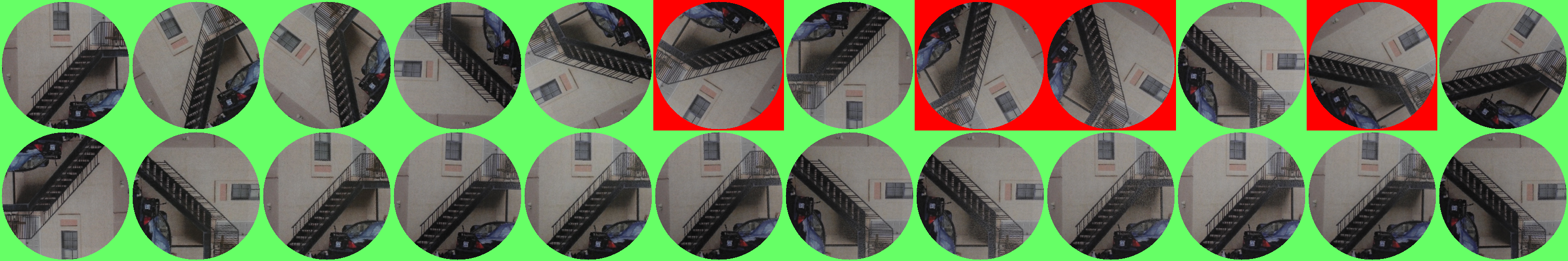}
\vspace{0.1cm}\\
\includegraphics[width=\linewidth]{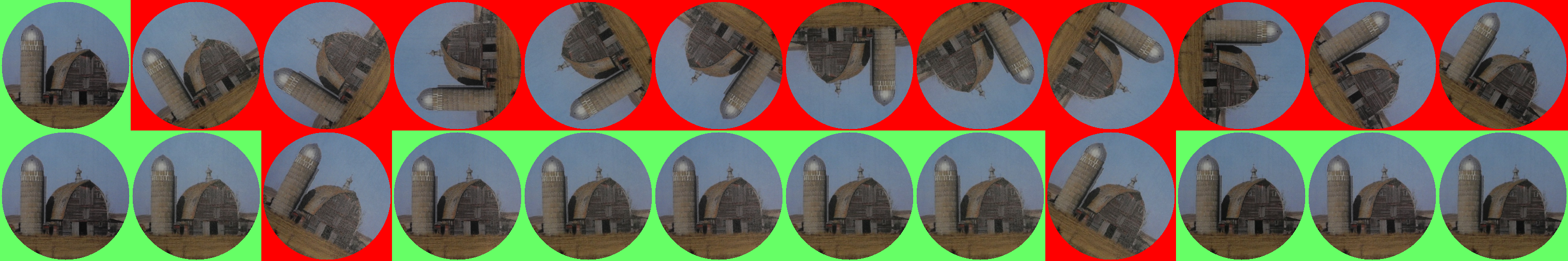}
\vspace{0.1cm}\\
\includegraphics[width=\linewidth]{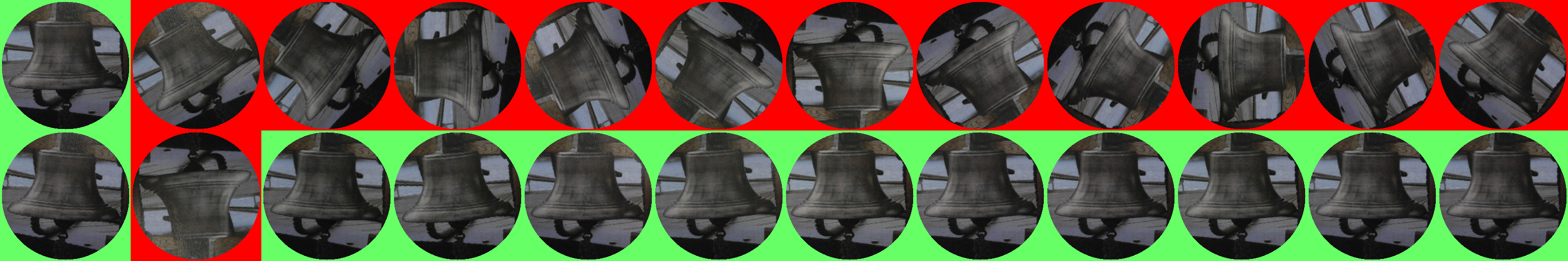}
\caption{All re-digitized photos raw (top rows) and after AMR (bottom rows) with their corresponding classifications color coded.}
\label{fig_all_sanity}
\end{figure}

Figure \ref{fig_printout} shows one example of the printouts that were used to create the manually rotated images for Section \ref{sec_sanity}. The additional helper lines were printed to facilitate the creation of photographs of the printout at exactly the twelve desired angles. Figure \ref{fig_all_sanity} shows the re-digitized images at an angle (top rows) and after AMR (bottom rows) with their classifications color coded analogous to Figure \ref{fig_sanity} but for all seven test images.



\end{document}